\documentclass{article}

\usepackage[numbers]{natbib}
\usepackage[preprint]{neurips_2024}

\usepackage[utf8]{inputenc} 
\usepackage[T1]{fontenc}    
\usepackage{url}            
\usepackage{booktabs}       
\usepackage{amsfonts}       
\usepackage{nicefrac}       
\usepackage{microtype}      
\usepackage{colortbl}
\usepackage{xcolor}         
\usepackage{multicol}
\usepackage{amsmath}
\usepackage{makecell} 
\usepackage{float}
\usepackage{pifont}       
\usepackage{bbding}       
\usepackage{fontawesome}  
\usepackage{wrapfig}
\usepackage{tcolorbox}
\usepackage{soul}
\usepackage{xspace}
\usepackage{subcaption}
\usepackage{wrapfig}

\definecolor{codeblue}{rgb}{0.25,0.5,0.5}
\definecolor{myblue}{rgb}{0.88,0.98,1}
\definecolor{mygreen}{rgb}{0.92, 1.0, 0.92}
\definecolor{myred}{rgb}{1, 0.9, 0.9}
\definecolor{mygray}{gray}{0.96}

\definecolor{Highlight}{HTML}{E8F8F5}
\definecolor{midgreen}{HTML}{69c5a3}
\definecolor{midblue}{HTML}{69a3f1}
\definecolor{darkgreen}{HTML}{38a3a5}
\definecolor{darkred}{HTML}{d00000}
\definecolor{darkblue}{HTML}{143b59}
\definecolor{mywhite}{HTML}{ffffff}

\definecolor{hotpink}{RGB}{59, 115, 227}
\usepackage[breaklinks,letterpaper,pdfpagelabels,pagebackref,colorlinks,allcolors=blue]{hyperref}
\usepackage{cleveref}

\title{Visual Instruction Tuning with Chain of Region-of-Interest }

\newcommand{\modelname}{CoRoI\xspace}

\author{Yixin Chen\thanks{Work done at Amazon} \\
  Amazon Web Services \\\And
  Shuai Zhang \\
  Amazon Web Services \\\AND
  Boran Han \\
  Amazon Web Services \\\And
  Bernie Wang \\
  Amazon Web Services
  }

\begin{document}

\maketitle

\begin{abstract}
High-resolution (HR) images are pivotal for enhancing the recognition and understanding capabilities of multimodal large language models (MLLMs). However, directly increasing image resolution can significantly escalate computational demands. In this study, we propose a method called Chain of Region-of-Interest (\textbf{\modelname}) for Visual Instruction Tuning, aimed at alleviating the computational burden associated with high-resolution images for MLLMs. Drawing inspiration from the selective nature of the human visual system, we recognize that not all regions within high-resolution images carry equal importance. \modelname seeks to identify and prioritize the most informative regions, thereby enhancing multimodal visual comprehension and recognition while circumventing the need for processing lengthy HR image tokens. Through extensive experiments on 11 benchmarks, we validate the efficacy of \modelname across varying sizes, ranging from 7B to 34B in parameters. Our models consistently demonstrate superior performance across diverse multimodal benchmarks and tasks. Notably, our method outperforms LLaVA-NeXT on almost all benchmarks and our finetuned 34B model surpasses proprietary methods like Gemini Pro 1.0 on six benchmarks, as well as outperforming GPT-4V on MMB, SEED-I, and MME. 





  
\end{abstract}

\section{Introduction} \label{sec:intro}



Multimodal Large Language Models (MLLMs)~\cite{team2023gemini, gpt4v, llava, llamaadapter, ofa, llava15, minigpt4, frozen, flamingo} have demonstrated outstanding performance across a variety of multimodal tasks. Despite this, their recognition and comprehension capabilities are limited by low-resolution image inputs, as these models are generally trained with fixed resolutions such as $336\times336$ or $448\times448$~\cite{llava15,llava,internvl15}. This inflexible design makes it challenging for them to identify small and blurry objects, capture crucial details, and generate fine-grained results.

Recent approaches have started incorporating high-resolution images to enhance MLLMs' performance~\cite{10246988,cogagent,wang2023cogvlm,llavanext,tong2024eyes,li2023otterhd}. For instance, LLaVA-Next~\cite{llavanext} extends LLaVA-1.5~\cite{llava15} by integrating high-resolution image tokens to grasp more visual details. InternLM-XComposer2-4KHD~\cite{internlmxcomposer24khd} uses compression techniques to accommodate more tokens for higher-resolution images. Additionally, LLaVA-HR~\cite{llavahr} utilizes mixture-of-resolution adapters to fuse low- and high-resolution. Other methods~\cite{cogagent,wang2023cogvlm} employ cross-attention layers to integrate high-resolution and low-resolution image tokens to avoid directly encoding high-resolution image tokens in LLM backbones. All these methods underscore the importance of high-resolution images in MLLMs.

However, existing solutions have several disadvantages, which we summarize as follows:  (1) Increasing resolution in a brute-force manner leads to a significant increase in token lengths (e.g.,  $5,329$ tokens for a $1,022 \times 1,022$ resolution image~\cite{llava15}), consequently resulting in quadratic growth in operations and computational demands for LLMs; (2) Commonly used vision encoders, such as pretrained ViTs~\cite{vit}, are typically trained on low-resolution images (e.g., $224 \times 224$), which can cause performance degradation when processing high-resolution images; (3) High-resolution image tokens cover all spatial locations indiscriminately, lacking focus on the areas relevant to the given queries. This can result in unnecessary computational overhead by processing regions that are irrelevant to the given questions; (4) Methods such as those in \cite{cogagent,wang2023cogvlm} can reduce the input tokens for LLMs, but they still require the cross-attention to be conducted between all high-resolution tokens and the low-resolution tokens. This may potentially result in many irrelevant or meaningless connections between tokens.
\begin{wrapfigure}{r}{0.5\textwidth} 
  \centering
  \includegraphics[width=0.5\textwidth]{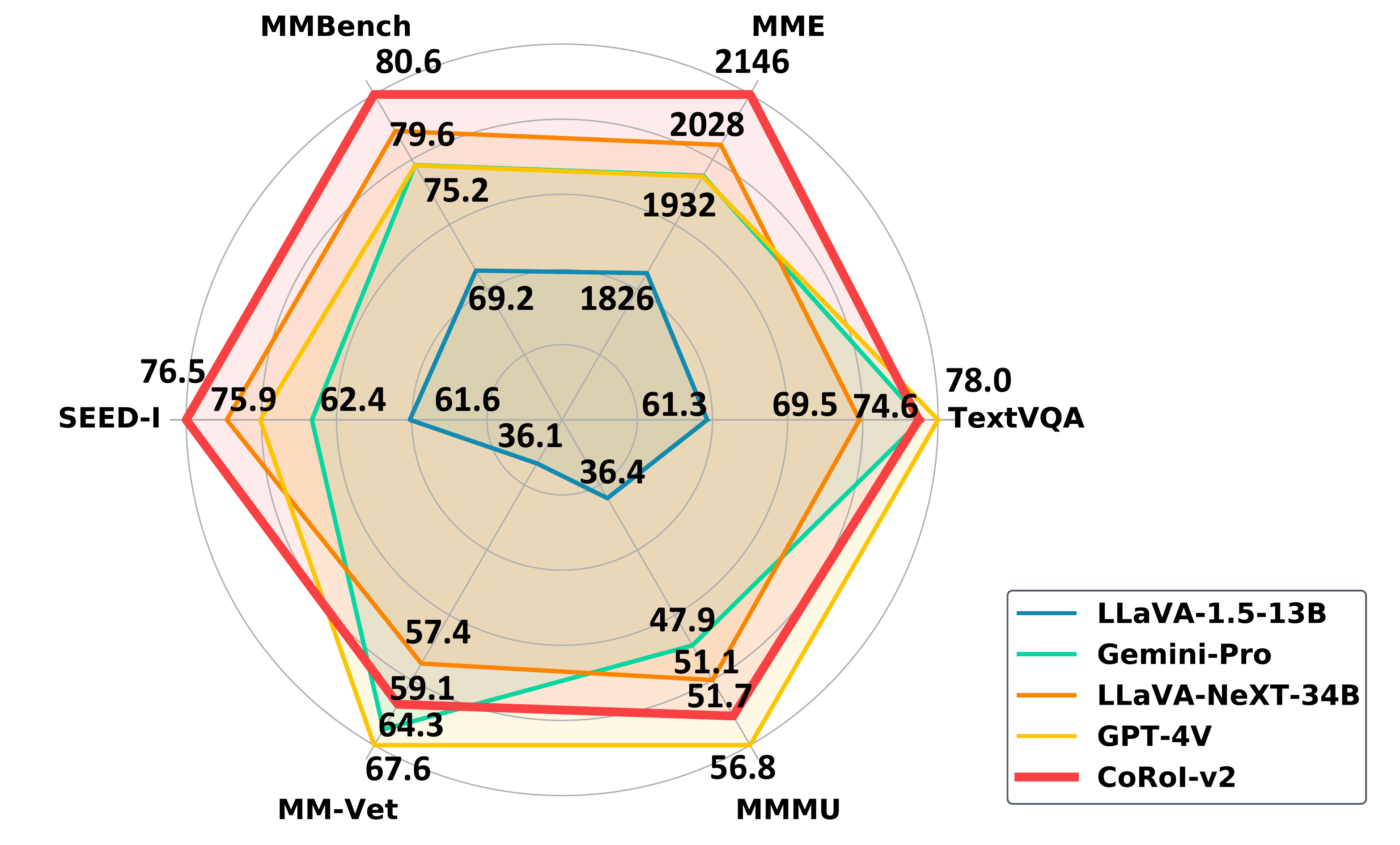}
  \caption{Zero-shot performance comparison with LLaVA-1.5, LLaVA-NeXT, Gemini-Pro 1.0, and GPT-4V.}
  \label{fig:your-label}
  \vspace{-1em}
\end{wrapfigure}



To address these issues, we propose a method named Chain of Region-of-Interest (\textbf{\modelname}) for visual instruction fine-tuning. \modelname consists of two main steps: (1) \textit{Chain of Region-of-Interest Extraction}: \modelname captures the most informative regions from high-resolution images by integrating visual clues (i.e., low-resolution (LR) image tokens) and language clues (i.e., text questions). Mimicking human eye movement, we design a region-of-interest movement mechanism that can generate a sequence of visually relevant regions from high-resolution images. This approach eliminates the need to process the entire high-resolution image, thereby reducing computational demands; (2) \textit{Incorporating Chain of RoI into LLMs}: \modelname injects the extracted regions of interest into the hidden layers of the LLMs, enabling cross-attention with the corresponding low-resolution image tokens.  This ensures a tight interaction between the high-resolution and low-resolution image tokens at different abstraction levels. To summarize, our contributions are:
\begin{itemize}
    \item We proposed \modelname, a visual instruction fine-tuning method by enforcing the model to only focus on relevant regions of high-resolution images, allowing the model to handle high-resolution images effectively and efficiently. 
    
    \item We applied \modelname to open-source LLMs with parameter sizes ranging from 7B to 34B and built three model variants, demonstrating that \modelname can consistently enhance multimodal recognition and comprehension capabilities.

    \item We conducted experiments on 11 benchmark datasets, including SEED-Image~\cite{seed}, MMBench~\cite{mmbench}, MME~\cite{mme}, MM-Vet~\cite{mmvet}, MMMU~\cite{mmmu}, MathVista~\cite{mathvista}, TextVQA~\cite{textvqa}, GQA~\cite{gqa}, VQA v2~\cite{vqadataset}, VizWiz~\cite{vizwiz}, and ScienceQA-Image~\cite{scienceqa}. Our results show that \modelname achieves state-of-the-art performance compared to open-source and proprietary models, consistently outperforming LLaVA-NeXT~\cite{llavanext} and Gemini Pro 1.0~\cite{team2023gemini}, occasionally matching or surpassing proprietary models like GPT-4V~\cite{gpt4v}.
    
    \item We conducted comprehensive model analyses, including ablation and case studies, to elucidate the inner workings and demonstrate the model's performance in real-world scenarios.
\end{itemize}




\section{Related Work}


\paragraph{Multimodal Large Language Models} 
The recent advancements in LLMs (e.g., ChatGPT~\cite{gpt3}, GPT-4~\cite{gpt4nov}, LLaMA~\cite{llama}, Claude~\cite{claude3}, and Mixtral~\cite{mixtral}) have sparked a growing interest in scaling up multimodal large language models (MLLMs). By integrating LLMs with vision encoders like ViT~\cite{vit} and CLIP~\cite{clip}, these models can handle various visual-language understanding tasks such as visual question answering and image captioning~\cite{llamaadapter, blip2, llava, llava15}. MLLMs mainly differ in their methods of combining LLMs with visual encoders. For example, some methods inject image tokens into the hidden layers of LLMs~\cite{llamaadapter, flamingo, openflamingo}, while others project them into the text token embedding space and treat them equally as text tokens~\cite{llava, llava15}. Additionally, it is possible to train MLLMs from scratch without using existing LLMs as backbones. These methods typically tokenize data from different modalities into the same embedding space and use them in an interleaved manner~\cite{ofa, perceiver, perceiverio, unival, uniperceiver, unifiedio, team2024chameleon} with causal language modeling training objectives.

\paragraph{MLLMs with High-Resolution Images} Existing MLLMs are typically trained with fixed resolutions, such as \(336 \times 336\) and \(448 \times 448\)~\cite{llava,llava15,llamaadapter, ofa, minigpt4,chen2024camml, frozen, flamingo}. This limitation makes it challenging for these models to perceive small or blurry objects in images, leading to failures in tasks such as OCR and document understanding, where high resolution is necessary to make details clear and recognizable and allow for finer distinctions between objects. Recently, many methods have introduced high-resolution inputs to enhance MLLMs' capability by providing more fine-grained visual features~\cite{10246988,cogagent,wang2023cogvlm,llavanext,tong2024eyes,li2023otterhd}. Examples include LLaVA-Next~\cite{llavanext}, LLaVA-HR~\cite{llavahr}, OtterHD~\cite{li2023otterhd}, InternLM-XComposer2-4KHD~\cite{internlmxcomposer24khd} and InternVL 1.5~\cite{internvl15}. Processing high-resolution images is a nontrivial task, particularly because most commonly used vision encoders (e.g., ViT) cannot handle high-resolution images directly. To avoid information loss, many methods~\cite{llava15,llavanext,cogagent} divide high-resolution images into smaller patches and encode them independently. These patches are then concatenated for further processing. While this approach can enhance performance, it also results in a quadratic increase in computational cost due to the long context length when working with LLMs. Differing from existing methods, \modelname can easily handle high-resolution images via the proposed chain of region-of-interest mechanism. This method extracts regions of interest from high-resolution images leveraging both language and visual signals. These identified regions of interest are then used to facilitate LLM generation. By focusing solely on the most informative regions, it circumvents the necessity to encode the entire high-resolution image. Moreover, the entire process is differentiable and does not require human intervention.


\section{Method}

\begin{figure}
    \centering
    \vspace{-10mm}
\includegraphics[width=1.0\linewidth]{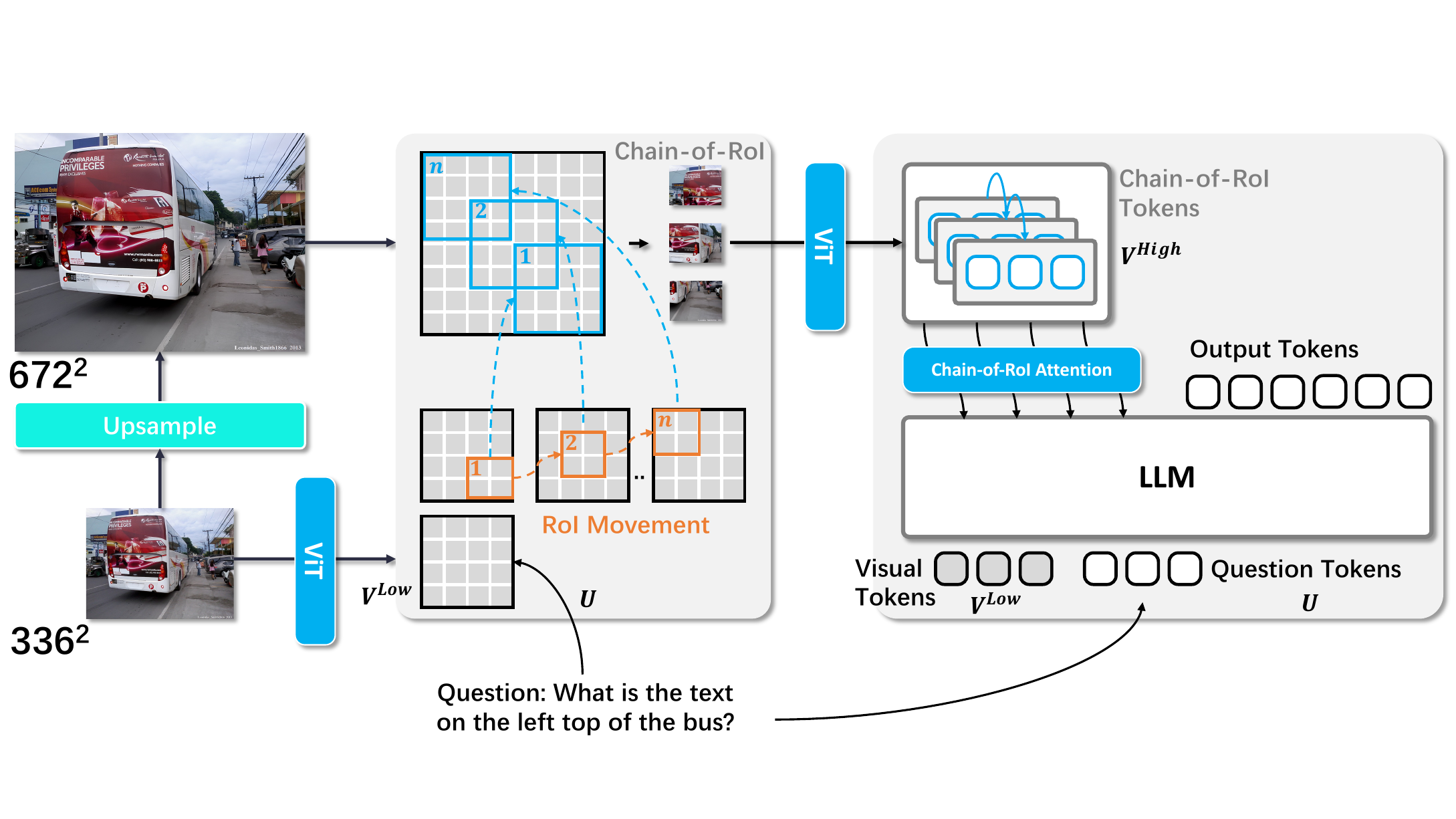}
    \vspace{-10mm}
    \caption{Illustration of \modelname. Low-resolution images and text clues are used to identify a chain of regions of interest in the corresponding high-resolution images. These regions are then injected into the hidden states of the LLMs to facilitate answer generation. Here, $m=1$, $\lambda=2$, $n=3$.}
    \label{fig:framework}
    \vspace{-1.5em}
\end{figure}

\begin{figure}[t]
    \centering
    \includegraphics[width=1.0\linewidth]{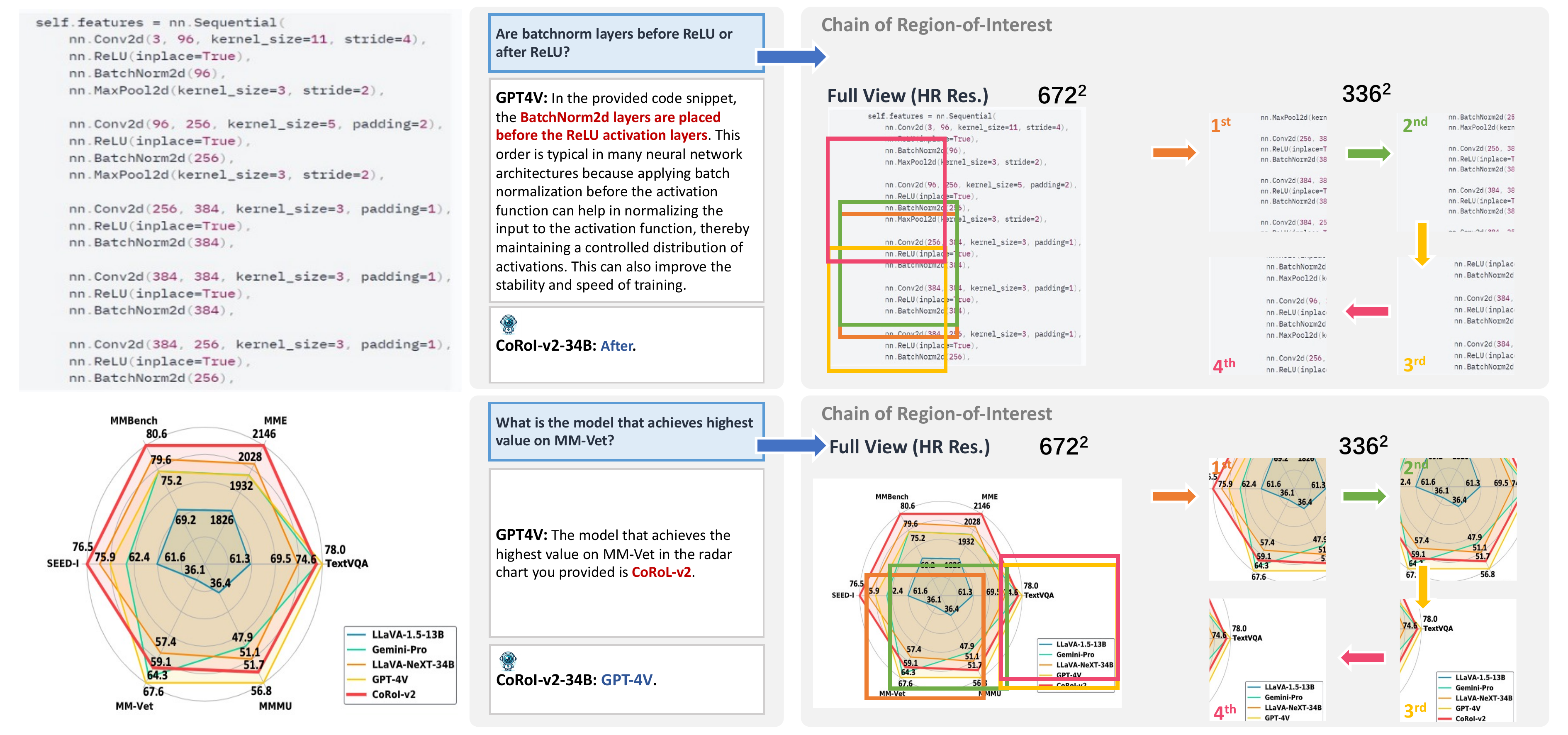}
    \caption{\modelname enables the model to extract a chain of region-of-interest during inference. We used variant: \modelname-v2 (Hermes-2-Yi-34B). The selected regions are closely relevant to the asked questions. We also provide the answers given by GPT-4V.}
    \label{fig:demo_compare}
    \vspace{-1em}
\end{figure}

Figure~\ref{fig:framework} illustrates the proposed framework. It operates by first upsampling the resolution of input low-resolution images to yield multiple high-resolution images with $m$ different scaling factors $\lambda \in \{2,3,4,...\}$. Subsequently, the low-resolution image and text questions are employed to pinpoint the most informative and relevant regions via the chain of region-of-interest extraction module. Once identified, corresponding regions are extracted from the high-resolution image as clips. These clips, with high-quality local information, are then injected into the hidden layers of the LLM to enhance the generation accuracy.

Before delving into the model details, let us examine two real-world examples in Figure \ref{fig:demo_compare}, using the model trained via \modelname. Remarkably, \modelname empowers the model to capture the most informative and relevant regions from the given high-resolution images, tailored to the asked questions.


\subsection{Chain of Region-of-Interest Extraction}
\begin{figure}[t!]
\centering
\begin{subfigure}[t]{0.48\linewidth}
\includegraphics[width=\linewidth]{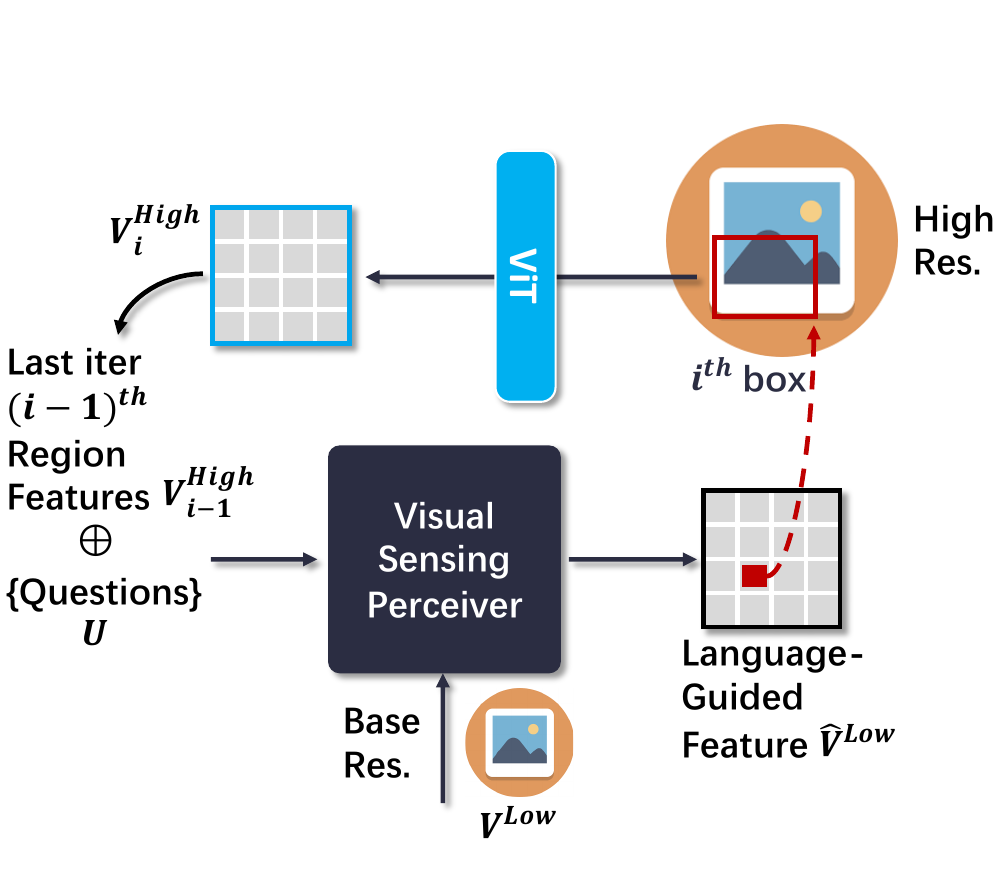}

\caption{Chain of Region-of-Interest Extraction.}
\label{fig:visual_sensing_module}
\end{subfigure}
\quad
\begin{subfigure}[t]{0.48\linewidth}
\includegraphics[width=\linewidth]{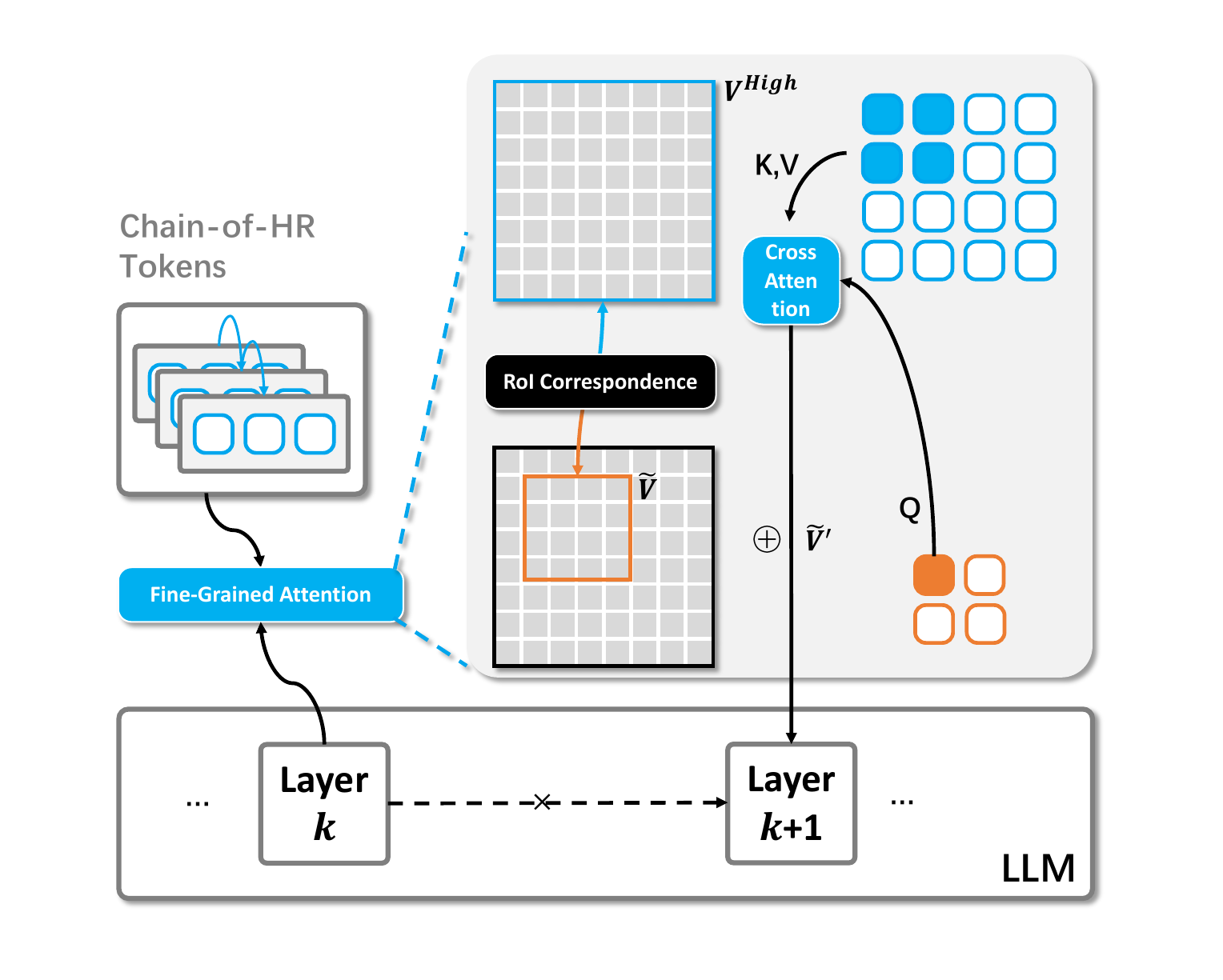}
\caption{Incorporating Chain of RoI into LLMs.}
\label{fig:attentive_module}
\end{subfigure}
\caption{
Modules design details.
}
\label{fig:module}
\vspace{-2.0em}
\end{figure}

Here, we introduce the chain of region-of-interest extraction methodology (shown in Figure \ref{fig:module}(a)).


\paragraph{Region-of-Interest Selection} To select the first high-resolution region, first, a cross-attention block is employed to process the low-resolution image feature, $\mathbf{V}^{Low} \in \mathbb{R}^{h \times w \times d}$ (extracted via ViT). This feature undergoes a self-attention layer to capture inherent visual relationships, followed by a cross-attention layer interacting with the question token embeddings $\mathbf{U}$. This integration enables the model to incorporate language clues. The results are then fed into a feed-forward layer, resulting in an output denoted as $\mathbf{\hat{V}}^{Low}:=f_{\theta^v}(\mathbf{V}^{Low}) \in \mathbb{R}^{h \times w \times 1}$. Next, to extract the most informative region, we employ a sliding window approach to traverse $\mathbf{\hat{V}}^{Low}$ and obtain a couple of candidate regions. The region with the highest pooled (average pooling) value is considered the current most informative region. Once we obtain the region location, we can extract the corresponding image clip from the high-resolution images. The sliding window is designed as follows: if the low-resolution image has dimensions of $336 \times 336$ and $\mathbf{\hat{V}}^{Low}$ is of size $24 \times 24$, and the high-resolution image has dimensions of $\lambda336 \times \lambda336$, the sliding window size would be $[24//\lambda, 24//\lambda]$. We set the stride size to $1$ and the padding size to $0$ by default. The extracted region clip from HR images will be dimension of $336 \times 336$.

\paragraph{Region-of-Interest Movement}
Inspired by how humans capture information with their eyes—\textit{focusing on one region of interest and then moving to the next for further clues}—we devised a region-of-interest movement mechanism. This mechanism operates iteratively, producing a sequence/chain of regions of interest. Initially, only the text question is used as the key/value for the RoI selection process (see paragraph above). After identifying the first region of interest from HR images, we encode this region clip using ViT to obtain a feature tensor $\mathbf{V}^{High}_{i}$, where $i = 0, \ldots, n$, represents the $i$-th iteration. We then concatenate $\mathbf{V}^{High}_{i}$ with the question token $\mathbf{U}$ and use the concatenated tensor as the key\&value for the \textit{cross-attention layer} mentioned above, allowing us to identify the second region of interest. As the process goes on, we will obtain $n$ regions of interest, forming a chain of regions of interest. If we have $m$ high-resolution images of different scaling factors, we will get $mn$ regions of interest in the end.


\subsection{Incorporating Chain of RoI into LLMs}
We integrate the extracted chain of regions of interest into LLMs (shown in Figure \ref{fig:module}(b)). Given that the number of regions, $mn$, can be large, we opt not to use them directly as input tokens for the LLM. Instead, we inject them into the LLM's hidden layers using a cross-attention module.


\paragraph{Cross Attention between RoI and LR Tokens}
In our framework, low-resolution tokens are directly used as the input for LLMs. To inject the $i^{\text{th}}$ region of interest into the $k^{\text{th}}$ layer of the LLM, we first identify the positions of the corresponding low-resolution tokens and extract the hidden states $\mathbf{\tilde{V}}^{Low}_{i, k}$ at layer $k$. Next, we perform cross-attention between the region of interest feature $ \mathbf{V}^{High}_{i}$ and $\mathbf{\tilde{V}}^{Low}_{i, k}$. The cross-attention module consists of a single linear layer and a layer normalization for the query, key, and value, respectively. Here, $\mathbf{\tilde{V}}^{Low}_{i, k}$ serves as the query, while $ \mathbf{V}^{High}_{i}$ acts as the key and value. 

\paragraph{Inject RoI into LLMs as Residual}The output, defined as $f_{\theta^{'}}(\mathbf{\tilde{V}}^{Low}_{i, k}, \mathbf{V}^{High}_{i})$, where $\theta^{'}$ is the parameters of cross-attention layer, is then used as a residual and added back to $\mathbf{\tilde{V}}^{Low}_{i, k}$. As such, the hidden state is modified to $\mathbf{\tilde{V}}^{Low}_{i, k} + f_{\theta^{'}}(\mathbf{\tilde{V}}^{Low}_{i, k}, \mathbf{V}^{High}_{i})$. This process injects high-resolution features into the LLMs with minimal additional cost. Layer normalization between LLM layers helps mitigate instability caused by the added residual.

\subsection{Model Training}
We have two training stages, with the vision encoder remaining frozen throughout. 

\textbf{1st stage - Projector Pretraining}: In this stage, we perform modality alignment between ViT and LLM using a trainable projector, following the approach in \cite{llava}, without incorporating the proposed modules. All other components except for the projector are frozen in this stage.

\textbf{2nd stage - Visual Instruction Fine-Tuning}: Here, we employ our modules and activate all parameters except the vision encoder. The model is trained by minimizing the causal language modeling loss: $\ell = -\sum_{i=1}^{|y|} \log p_{\theta} (y_i |\hat{y}_{1:i-1}, q)$, where \(\theta \leftarrow (\theta^{\text{LLM}}, \theta^{v}, \theta^{'})\) represents the model's trainable parameters (\(\theta^{\text{LLM}}\) being the LLM's parameters), \(y_i\) is the ground-truth target, and \(\hat{y}_{1:i-1}\) denotes the \(i-1\) preceding tokens of the output \(y_i\).


\section{Experiment}\label{sec:experiments}

\subsection{Experimental Setup}

\paragraph{Implementation Details}
We resized all raw training images to \( 336 \times 336 \) pixels, treating this as the low/base resolution. We then used bilinear interpolation to generate the corresponding high-resolution images. By setting \( m = 3 \) and \(\lambda\) to 2, 3, and 4, the resulting high-resolution images have dimensions of \( 672 \times 672 \), \( 1008 \times 1008 \), and \( 1344 \times 1344 \) pixels, respectively. The number of RoI ($n$) in each chain is set to $4$. We employed a CLIP-pretrained ViT-L~\cite{clip} model as the vision encoder, which remained frozen throughout the training process. All the training processes were conducted for one epoch using the AdamW optimizer and a cosine learning rate schedule, without further tuning. The initial learning rate was set to \( 1 \times 10^{-3} \) for projector pretraining and \( 2 \times 10^{-5} \) for visual instruction fine-tuning for all models, except for Hermes-2-Yi-34B~\cite{hermesyi}, where the instruction fine-tuning learning rate was set to \( 1 \times 10^{-5} \). All experiments were performed on 8 H100 GPUs with an accumulative batch size of 128, initialized using the DeepSpeed Zero3~\cite{rasley2020deepspeed} configuration. 

\paragraph{Model Variants} We used Mistral-7B~\cite{mistral} and Hermes-2-Yi-34B~\cite{hermesyi} as the LLM backbones. For each backbone, we created two model versions by varying the input tokens. In the first version (denoted as \modelname-v1), we used \( 336 \times 336 \) images, corresponding to 576 tokens as the LLM's input. In the second version (denoted as \modelname-v2), we also used \( 672 \times 672 \) images as input for the LLMs, resulting in a total of 2880 tokens (576 tokens from the \( 336 \times 336 \) images plus 2304 tokens from the \( 672 \times 672 \) images). In this case, the RoI features were only injected into the tokens from the \( 336 \times 336 \) images. All in all, the LLM backbone does not have to deal with a very long context length.

\begin{table}[t!]
\centering
\caption{Comparison with state-of-the-art methods on comprehensive zero-shot benchmarks. Res.$^{*}$ means the visual resolution of LLM image input tokens.}
\scalebox{0.78}{
\begin{tabular}{lll|cccccc}
\toprule
Method & LLM & Res.$^{*}$ & {\bf SEED$^{\text{I}}$} & {\bf MMB} & {\bf MME} & {\bf MM-Vet} & {\bf MMMU}$_{v}$  & {\bf MathVista}   \\
\midrule
\ MobileVLM-V2~\cite{mobilevlmv2} & Vicuna-7B & 336$^2$ & --   & 69.2 & 1560 & -- & -- & --   \\
\ InstructBLIP~\cite{instructblip} & Vicuna-7B & 224$^2$ & 58.8 & 36.0 & -- & 26.2 & -- & 25.3  \\
\ InstructBLIP~\cite{instructblip} & Vicuna-13B & 224$^2$ & --  & -- & 1213 & 25.6 & --  & --     \\
\ Qwen-VL~\cite{qwenvl} & Qwen-7B & 448$^2$ &  62.3 & 38.2 & -- & -- & --  & --  \\
\ Qwen-VL-Chat~\cite{qwenvl} & Qwen-7B & 448$^2$ & 65.4  & 60.6 & 1488  & -- & 35.9 &  --   \\
\ Shikra~\cite{shikra} & Vicuna-13B & 224$^2$ & --  & 58.8 & --  & -- & -- & -- \\
\ IDEFICS-80B~\cite{obelics} & LLaMA-65B & 224$^2$ & 53.2 & 54.5 & --  & -- & -- & --  \\
\ LLaVA-1.5~\cite{llava15} & Vicuna-7B & 336$^2$ & --  & 65.2 & 1511  & 31.1 & -- & --   \\
\ LLaVA-1.5~\cite{llava15} & Vicuna-13B & 336$^2$ & 68.2  & 69.2 & 1531/295  & 36.1 & 36.4 & 27.6   \\

\midrule
\rowcolor{mygray}
\rowcolor{mygray}
\ \modelname-v1 & Mistral-7B & 336$^2$ & 72.0 & 69.0 & 1528/328  & 44.5 & 36.8 &  36.2 \\
\midrule
\ OtterHD~\cite{li2023otterhd} & Fuyu-8B & 1024$^2$ &  -- & 53.6 & 1314  & -- & -- & --   \\
\ CogVLM-Chat~\cite{wang2023cogvlm} & Vicuna-7B & 490$^2$ & 72.5  & 63.7 & --  & 51.1 & 41.1 & 34.5  \\
\ LLaVA-HR-X~\cite{llavahr} & Vicuna-13B & 1024$^2$ & -- & -- & 1487 & 35.5 & -- & -- \\ 
\ LLaVA-NeXT~\cite{llavanext} & Vicuna-7B & 672$^2$ & 70.2   & 68.1 & 1519/332  & 43.9 & 35.8 & 34.6   \\
\ LLaVA-NeXT~\cite{llavanext} & Vicuna-13B & 672$^2$ & 71.9 & 70.7 & 1575/326 & 48.4 & 36.2 & 35.3    \\
\ LLaVA-NeXT~\cite{llavanext} & Hermes-2-Yi-34B & 672$^2$ & \underline{75.9}  & \underline{79.6} & 1631/397  & 57.4 & 51.1 &   46.5   \\
\midrule
\rowcolor{mygray}
\rowcolor{mygray}
\ \modelname-v2 & Mistral-7B & 672$^2$ & 73.1 & 69.9  & 1540/339 & 49.6 & 36.9  & 37.5  \\
\rowcolor{mygray}
\ \modelname-v2 & Hermes-2-Yi-34B & 672$^2$ & \textbf{76.5} & \textbf{80.6}  &  \underline{1662/484} &  59.1  & \underline{51.7}  &    \underline{47.2} \\
\midrule
\ Gemini Pro 1.0~\cite{team2023gemini} & Private & --  & 62.4  & 75.2 & 1496/436  & 64.3 & 47.9 & 45.2   \\
\ Qwen-VL-Plus~\cite{qwenvl} & Private & -- & 72.7  & 66.2 & \textbf{1681/502} & -- & 45.2 &  43.3  \\
\ GPT-4V~\cite{gpt4v} & Private & -- & 69.1 & 75.1 & 1409/517  & \textbf{67.6} & \textbf{56.8} &  \textbf{49.9}    \\
\bottomrule
\end{tabular}}
\vspace{-6mm}
\label{tab:main_result_bench}
\end{table}

\begin{table}[t]
\centering
\caption{Comparison with state-of-the-art methods on Vision-Language tasks. Res.$^{*}$ means the visual resolution of LLM image input tokens.}
\scalebox{0.8}{
\begin{tabular}{lll|ccccc}
\toprule
Method & LLM & Res.$^{*}$ & {\bf TextVQA (VQA}$^\text{T}$) & {\bf GQA} & {\bf VQAv2} & {\bf Vizwiz} & \bf SQA$^\text{I}$   \\
\midrule
\ MobileVLM-V2\cite{mobilevlmv2} & Vicuna 7B & 336$^2$ & 62.3  & 62.6 & -- & -- & 74.8  \\
\ InstructBLIP~\cite{instructblip} & Vicuna-7B & 224$^2$ & 50.1 & 49.2 & -- & 34.5 & 60.5   \\
\ InstructBLIP~\cite{instructblip} & Vicuna-13B & 224$^2$ & 50.7 & 49.5 & -- & 33.4 & 63.1  \\
\ Qwen-VL~\cite{qwenvl} & Qwen-7B & 448$^2$ & 63.8  & 59.3 & 78.8 & 35.2 & 67.1   \\
\ Qwen-VL-Chat~\cite{qwenvl} & Qwen-7B & 448$^2$ & 61.5  & 57.5 & 78.2 & 38.9 & 68.2    \\
\ Shikra~\cite{shikra} & Vicuna-13B & 224$^2$ & --  & -- & 77.4 & -- & --  \\
\ IDEFICS-80B~\cite{obelics} & LLaMA-65B & 224$^2$ & 30.9  & 45.2 & 60.0 & 36.0 & --   \\
\ LLaVA-1.5~\cite{llava15} & Vicuna-7B & 336$^2$ & 58.2  & 62.0 & 78.5 & 50.0 & 66.8   \\
\ LLaVA-1.5~\cite{llava15} & Vicuna-13B & 336$^2$ & 61.3  & 63.3 & 80.0 & 53.6 &  71.6 \\
\midrule
\rowcolor{mygray}
\rowcolor{mygray}
\ \modelname-v1 & Mistral-7B & 336$^2$ & 64.8 & 64.5 & 81.2 & 56.7 & 73.1   \\
\midrule
\ CogVLM-Chat~\cite{wang2023cogvlm} & Vicuna-7B & 490$^2$ & 70.4  & -- & 82.3 & -- & --   \\
\ LLaVA-HR-X~\cite{llavahr} & Vicuna-13B & 1024$^2$ & \underline{70.9} & 65.2 & 82.6 & 56.6 & 68.0 \\
\ LLaVA-NeXT~\cite{llavanext} & Vicuna-7B & 672$^2$ & 64.9  & 64.2 & 81.8 & 57.6 & 70.1    \\
\ LLaVA-NeXT~\cite{llavanext} & Vicuna-13B & 672$^2$ & 67.1 & 65.4 & 82.8 & 60.5 & 73.6 \\
\ LLaVA-NeXT~\cite{llavanext} & Hermes-2-Yi-34B & 672$^2$ & 69.5  & \underline{67.1} & \underline{83.7} & \textbf{63.8} & \underline{81.8}    \\
\midrule
\rowcolor{mygray}
\rowcolor{mygray}
\ \modelname-v2 & Mistral-7B & 672$^2$ & 69.7 & 65.1 & 82.7 & 59.6 & 73.4     \\
\rowcolor{mygray}
\ \modelname-v2 & Hermes-2-Yi-34B & 672$^2$ & \textbf{74.0} & \textbf{67.8} & \textbf{84.0}& \underline{63.2} & \textbf{82.8}    \\
\bottomrule
\end{tabular}}
\label{tab:main_result_task}
\vspace{-3mm}
\end{table}

\paragraph{Datasets}
For model optimization, we create high-quality data to enhance cross-modality understanding and generation. This dataset includes LLaVA-558K and ALLaVA-Caption-4V~\cite{allava} ($\sim$1.2 million caption pairs) for modality alignment during projector pretraining and LLaVA-665K (made up of LLaVA-Instruct-158K~\cite{llava}, ShareGPT-40K~\cite{sharegpt}, VQAv2~\cite{vqadataset}, GQA~\cite{gqa},
OKVQA~\cite{okvqa}, OCRVQA~\cite{ocrvqa}, A-OKVQA~\cite{aokvqa}, RefCOCO~\cite{refcoco} and VG~\cite{vg}) without TextCaps~\cite{textcap}, ShareGPT4V~\cite{sharegpt4v}, LAION-GPT4V~\cite{laiongpt4v}, DocVQA~\cite{docvqa}, AI2D~\cite{ai2d}, ChartQA~\cite{chartqa}, DVQA~\cite{dvqa} and ALLaVA-Instruct-4V~\cite{allava} ($\sim$1.5 million single- or multi-round conversations) for visual instruction fine-tuning. We evaluated our models on widely-adopted zero-shot multimodal benchmarks, including SEED~\cite{seed} (Image), MMB~\cite{mmbench} (MMBench), MME~\cite{mme}, MM-Vet~\cite{mmvet}, MMMU~\cite{mmmu}, and MathVista~\cite{mathvista} datasets. Additionally, we reported results on well-known visual question answering datasets, such as VQA$^{\text{T}}$ (TextVQA), GQA~\cite{gqa}, VQA v2~\cite{vqadataset}, VizWiz~\cite{vizwiz}, and SQA$^{\text{I}}$~\cite{scienceqa} (ScienceQA-Image).


\subsection{Model Quantitative Analysis}

\subsubsection{Multimodal Instruction-Following}

\paragraph{Comprehensive Multimodal Benchmarks}

In Table~\ref{tab:main_result_bench}, we compare our approach with previous leading open-source and closed-source methods across various comprehensive zero-shot multimodal benchmarks. These benchmarks assess the model's visual understanding, reasoning, multidisciplinary abilities, and even logical thinking and math capabilities. Overall, we observe a consistent performance boost from the three variants over open-source models with the same parameter size and input resolutions, confirming the effectiveness of our proposed method. Our most performant variant, \modelname-v2 (Hermes-2-Yi-34B), surpasses proprietary closed-source models such as Gemini Pro 1.0~\cite{team2023gemini} on 5/6 benchmarks, and outperforms Qwen-VL-Plus~\cite{qwenvl} in most cases. It also occasionally performs better than or is on par with GPT-4V~\cite{gpt4v} on all the benchmarks.


\paragraph{Visual Question Answering Benchmarks}In Table \ref{tab:main_result_task}, we also present a comparison of \modelname with existing methods on widely used visual question answering benchmarks. Datasets such as TextVQA (VQA$^{\text{T}}$) require the model to have certain OCR (Optical Character Recognition) capabilities to read and reason over the text and scene in the given images. Similarly, noticeable performance improvements can be observed on the five datasets compared to baselines with the same LLM backbones. In particular, the performance increase using \modelname on VQA$^{\text{T}}$ is considerably significant, demonstrating its ability to handle distinct details from images by incorporating regions of interest from high-resolution images.


\begin{table}[t]
\setlength{\tabcolsep}{6pt}
 \centering
  \caption{Ablation on model design. In upper part, \textcolor{darkgreen}{\rule{4.0mm}{2mm}} is the improvement against LLaVA-1.5, and in lower part \textcolor{darkred}{\rule{4.0mm}{2mm}} indicates performance degradation compared with \modelname-v1.
 }
 \scalebox{0.72}{
\begin{tabular}{lccc|cr@{\hspace{3pt}}l|cr@{\hspace{3pt}}l|c r@{\hspace{3pt}}l}
  \toprule
  Method &  Res. & HR Res. & LLM Tokens & \multicolumn{3}{c|}{\bf TextVQA (VQA$^\text{T}$)} & \multicolumn{3}{c|}{\bf MMBench} & \multicolumn{3}{c}{\bf MME} \\
  \midrule
  \midrule
    \ LLaVA-1.5  & 336$^2$ & -- & 576 & 58.2 & & & 65.2 & & & 1510 & & \\
    \midrule
  \textit{\textbf{Apply \modelname:}} & & & & & & & & & & &  \\
  \ \modelname-v1   & 336$^2$ & 2/3/4$\times$ & 576 & 63.3 & \textcolor{darkgreen}{+5.1} & \textcolor{mywhite}{\rule{1.80mm}{2mm}}\textcolor{darkgreen}{\rule{5.10mm}{2mm}} & 69.0 & \textcolor{darkgreen}{+3.8} & \textcolor{mywhite}{\rule{1.60mm}{2mm}}\textcolor{darkgreen}{\rule{3.80mm}{2mm}}  & 1524 & \textcolor{darkgreen}{+14} & \textcolor{mywhite}{\rule{5.15mm}{2mm}}\textcolor{darkgreen}{\rule{0.7mm}{2mm}} \\ 
  \ {\em w/o} Visual Movement   & 336$^2$ & 2/3/4$\times$ & 576 & 62.5 & \textcolor{darkgreen}{+4.3} & \textcolor{mywhite}{\rule{1.80mm}{2mm}}\textcolor{darkgreen}{\rule{4.30mm}{2mm}} & 68.4 & \textcolor{darkgreen}{+3.2} & \textcolor{mywhite}{\rule{1.60mm}{2mm}}\textcolor{darkgreen}{\rule{3.20mm}{2mm}} & 1513 & \textcolor{darkgreen}{+3} & \textcolor{mywhite}{\rule{5.15mm}{2mm}}\textcolor{darkgreen}{\rule{0.15mm}{2mm}} \\
  \midrule
  \textit{\textbf{Layers Location ($k$):}} & & & & & & & & & & &  &  \\
  \ {\em + \{15\}$^{th}$ layer}  & 336$^2$ & 2/3/4$\times$ & 576 & 62.6 & \textcolor{darkred}{-0.7} & \textcolor{mywhite}{\rule{1.1mm}{2mm}}\textcolor{darkred}{\rule{0.7mm}{2mm}} & 68.0 &  \textcolor{darkred}{-1.0} & \textcolor{mywhite}{\rule{0.6mm}{2mm}}\textcolor{darkred}{\rule{1.0mm}{2mm}} & 1423 & \textcolor{darkred}{-101} & \textcolor{mywhite}{\rule{0.1mm}{2mm}}\textcolor{darkred}{\rule{5.05mm}{2mm}} \\
  \ {\em + \{7, 15, 23\}$^{th}$ layer}   & 336$^2$ & 2/3/4$\times$ & 576 & 63.0 &\textcolor{darkred}{-0.3} & \textcolor{mywhite}{\rule{1.5mm}{2mm}}\textcolor{darkred}{\rule{0.3mm}{2mm}} & 68.4 & \textcolor{darkred}{-0.6} & \textcolor{mywhite}{\rule{1.0mm}{2mm}}\textcolor{darkred}{\rule{0.6mm}{2mm}} & 1431  & \textcolor{darkred}{-93} & \textcolor{mywhite}{\rule{0.5mm}{2mm}}\textcolor{darkred}{\rule{4.65mm}{2mm}} \\ 
  \textit{\textbf{RoI Nums ($n$):}} & & & & & & & & & & & &  \\
  \ {+ 1 RoIs}   & 336$^2$ & 2/3/4$\times$ & 576 & 62.4 & \textcolor{darkred}{-0.9} & \textcolor{mywhite}{\rule{0.9mm}{2mm}}\textcolor{darkred}{\rule{0.9mm}{2mm}} & 68.5 & \textcolor{darkred}{-0.5} & \textcolor{mywhite}{\rule{1.1mm}{2mm}}\textcolor{darkred}{\rule{0.5mm}{2mm}} &  1421  & \textcolor{darkred}{-103} & \textcolor{darkred}{\rule{5.15mm}{2mm}} \\ 
  \ {+ 2 RoIs}   & 336$^2$ & 2/3/4$\times$ & 576 & 61.9 & \textcolor{darkred}{-1.4} & \textcolor{mywhite}{\rule{0.4mm}{2mm}}\textcolor{darkred}{\rule{1.4mm}{2mm}} & 68.7 & \textcolor{darkred}{-0.3} & \textcolor{mywhite}{\rule{1.3mm}{2mm}}\textcolor{darkred}{\rule{0.3mm}{2mm}} &  1434 &  \textcolor{darkred}{-90} & \textcolor{mywhite}{\rule{0.65mm}{2mm}}\textcolor{darkred}{\rule{4.5mm}{2mm}}  \\ 
  \textit{\textbf{Multi-scale HR Resolution ($m$, $\lambda$):}} & & & & & & & & & & & &  \\
    \ {+ 2$\times$}   & 336$^2$ & 2$\times$ & 576 & 61.5 & \textcolor{darkred}{-1.8} & \textcolor{darkred}{\rule{1.80mm}{2mm}} &  
67.4 & \textcolor{darkred}{-1.6} & \textcolor{darkred}{\rule{1.6mm}{2mm}} & 1486 &  \textcolor{darkred}{-38} & \textcolor{mywhite}{\rule{3.25mm}{2mm}}\textcolor{darkred}{\rule{1.9mm}{2mm}} \\ 
  \ {+ 3$\times$}   & 336$^2$ & 3$\times$ & 576  & 62.2 & \textcolor{darkred}{-1.1} & \textcolor{mywhite}{\rule{0.7mm}{2mm}}\textcolor{darkred}{\rule{1.1mm}{2mm}} & 68.3 & \textcolor{darkred}{-0.7} & \textcolor{mywhite}{\rule{0.9mm}{2mm}}\textcolor{darkred}{\rule{0.7mm}{2mm}} & 1454 & \textcolor{darkred}{-70} & \textcolor{mywhite}{\rule{1.65mm}{2mm}}\textcolor{darkred}{\rule{3.5mm}{2mm}} \\ 
  \ {+ 4$\times$}   & 336$^2$ & 4$\times$ & 576 & 62.7 & \textcolor{darkred}{-0.6} & \textcolor{mywhite}{\rule{1.2mm}{2mm}}\textcolor{darkred}{\rule{0.6mm}{2mm}} & 68.1 & \textcolor{darkred}{-0.9} & \textcolor{mywhite}{\rule{0.7mm}{2mm}}\textcolor{darkred}{\rule{0.9mm}{2mm}}  & 1473 & \textcolor{darkred}{-51} & \textcolor{mywhite}{\rule{2.6mm}{2mm}}\textcolor{darkred}{\rule{2.55mm}{2mm}} \\ 
  \ {+ 2/3$\times$}   & 336$^2$ & 2/3$\times$ & 576 & 62.1 & \textcolor{darkred}{-1.2} & \textcolor{mywhite}{\rule{0.7mm}{2mm}}\textcolor{darkred}{\rule{1.2mm}{2mm}} & 68.5 & \textcolor{darkred}{-0.5} & \textcolor{mywhite}{\rule{1.1mm}{2mm}}\textcolor{darkred}{\rule{0.5mm}{2mm}}  & 1487 & \textcolor{darkred}{-37} & \textcolor{mywhite}{\rule{3.3mm}{2mm}}\textcolor{darkred}{\rule{1.85mm}{2mm}} \\ 
  \ {+ 3/4$\times$}   & 336$^2$ & 3/4$\times$ & 576 & 62.0 & \textcolor{darkred}{-1.3} & \textcolor{mywhite}{\rule{0.5mm}{2mm}}\textcolor{darkred}{\rule{1.3mm}{2mm}} & 68.6 & \textcolor{darkred}{-0.4} & \textcolor{mywhite}{\rule{1.2mm}{2mm}}\textcolor{darkred}{\rule{0.4mm}{2mm}} & 1496 & \textcolor{darkred}{-28} & \textcolor{mywhite}{\rule{3.75mm}{2mm}}\textcolor{darkred}{\rule{1.4mm}{2mm}} \\ 
  \bottomrule
  \vspace{-10mm}
\end{tabular}}

 \label{tab:ablation_model_design}
\end{table}

\subsubsection{Component Analysis}

\paragraph{Model Design}

We will begin by examining the design choices of the proposed method and presenting the results in Table~\ref{tab:ablation_model_design}.
It is evident that the model demonstrates substantial improvements when our method is employed to enhance the original LLM baseline.
For example, with LR and HR set to 336 and 672 respectively, the model achieves increase in the TextVQA and MME datasets.
The addition of layers and positioning the attentive layers around the middle of the LLM further amplifies the performance, resulting in improvement in TextVQA compared to a single layer.
These findings highlight the significant impact of our method in leveraging more detailed visual cues.
Furthermore, expanding the RoI numbers from 1 to 4 continues to yield additional improvements with our method.
For instance, with 4 RoIs, it outperforms the baseline on the TextVQA~\cite{textvqa}, MME~\cite{mme}, and MM-Vet~\cite{mmvet} datasets, respectively.
The utilization of multi-scale high resolution images proves to be more advantageous than single-scale high resolution images. Using 2/3/4$\times$ high resolution visual features are notably superior to the 2$\times$ or 3$\times$ counterparts, demonstrating the effectiveness of the designed modules with scaled-up input resolutions.
\begin{wraptable}{r}{0.6\textwidth}
\vspace{-0.5em}
\setlength{\tabcolsep}{6pt}
 \centering
  \caption{Ablation on instruction-finetuning dataset, LLM input tokens: evaluated on VQA$^{\text{T}}$(TextVQA), MMBench and SEED benchmarks.
 }
 \scalebox{0.57}{
\begin{tabular}{lcc|c|c|c}
  \toprule
  Method &  Res. &  LLM Tokens & \multicolumn{1}{c|}{\bf VQA$^{\text{T}}$} & \multicolumn{1}{c|}{\bf MMBench} & \multicolumn{1}{c}{\bf SEED} \\
  \midrule
  \midrule
  LLaVA-1.5   & 336$^2$  & 576 & 58.2 & 68.0 & 58.6 \\
  
  \modelname-v1   & 336$^2$  & 576 & 63.3 & 69.0 & 60.1 \\ 
  \ {\em + ShareGPT4V}   & 336$^2$  & 576 & 64.6 & 68.7 & 60.8 \\ 
  \ {\em -- TextCaps}   & 336$^2$ &  576 & 59.3 & 68.4 & 60.4  \\ 
  \ {\em + LAION-GPT4V}   & 336$^2$ &  576 & 61.2 & 68.4 & 61.0 \\ 
  \ {\em + DocVQA,AI2D,ChartQA,DVQA}   & 336$^2$ &  576 & 60.8 & 68.8 & 61.8 \\ 
  \ {\em + ALLaVA}   & 336$^2$ &  576 & 63.2 & 69.2 & 62.6  \\ 
  \ {\em + 2x Res. Token}   & 336$^2$+672$^2$ &  2880 & 69.2 & 69.5 & 63.0 \\ 

  \bottomrule
\end{tabular}}
 \label{tab:ablation_dataset_scale}
\end{wraptable}

\vspace{-2em}
\paragraph{Dataset}

The importance of high-quality data cannot be emphasized enough for MLLMs. Our thorough analysis of the combined effects of different types of data, as shown in Table~\ref{tab:ablation_dataset_scale}, starts with a base model that incorporates our method trained on the LLaVA-665K~\cite{llava15} instruction tuning data. By integrating high-quality captions from ShareGPT-4V~\cite{sharegpt4v}, we observed improved visual alignment and performance gains. We tested the model's zero-shot performance on the TextVQA~\cite{textvqa} benchmark, specifically by removing TextCaps~\cite{textcap} data from the training set as previous studies have done. This adjustment resulted in a significant decrease in performance, highlighting the importance of specific types of data in the training process. To address this decline, we introduced additional high-quality captions from LAION-GPT-4V~\cite{laiongpt4v} and OCR-specific data such as DocVQA~\cite{docvqa} and AI2D~\cite{ai2d} \emph{etc.}, similar to the approach taken by~\cite{llava15}, thereby enhancing the model's OCR reasoning capabilities. By leveraging high-quality GPT-4V-generated responses from the ALLaVA~\cite{allava} dataset, our framework was able to improve the baseline respectively on TextVQA and MMBench datasets. This comprehensive evaluation emphasizes the crucial role of strategic integration of high-quality data in maximizing the potential of our framework.

\subsection{Model Qualitative Analysis}

\paragraph{Visualization of the Selected Chain of Region-of-Interest}

As seen in Figure~\ref{fig:demo_compare}, \modelname provides concise and correct answers in these two examples, whereas GPT-4V struggles to understand the visual spatial relationships or tends to produce hallucinated answers. On the right-hand side, we show the selected chain of regions of interest during model inference. It is evident that these regions can move reasonably over the high-resolution images and typically focus on the most informative areas. For example, in the code snippet example, the regions are centered around the ReLU and BatchNorm layers; in the Titanic example, the focused regions are crucial for answering the text questions accurately. This justifies the efficacy of integrating language clues in the RoI selection process.



\section{Limitations}\label{sec:limitations}

The highest resolution adopted in \modelname is $1344 \times 1344$, which may limit our model's capability. However, this limitation primarily stems from a shortage of computational resources. Additionally, we employ interpolation methods for upsampling to obtain high-resolution images, but this process may result in the loss of some key information. Nevertheless, this issue mainly arises due to the lack of publicly available large-scale and high-quality high-resolution training data.

\section{Conclusion}\label{sec:conclusion}
In this work, we introduce \modelname, a visual instruction fine-tuning method utilizing the proposed Chain of Region-of-Interest approach. \modelname aims to identify and prioritize the most informative regions in high-resolution images, thereby enhancing multimodal visual comprehension and recognition. We developed three model variants using \modelname, ranging from 7B to 34B parameters. Extensive experiments on 11 benchmarks validate the efficacy of \modelname across varying model sizes. We compared our models with both open-source models, such as LLaVA-NeXT, and closed-source models, such as Gemini Pro 1.0 and GPT-4V, demonstrating the superiority of our approach. Additionally, we conducted extensive model analyses to understand the inner mechanisms and demonstrate its capability in solving real-world multimodal tasks.



\clearpage



{\small
\bibliographystyle{ieee_fullname}
\bibliography{allbib}
}
\clearpage
\newpage
\appendix

\section{Appendix}

\paragraph{Experimental Details}
We present the experimental details of our model here. The default layers location $k$ is determined by first uniformly dividing all LLM layers into 4 sections, then doubling the layers within the section close to the middle of LLM. For example, using 32 layers Mistral-7B, we initially select \{7, 15, 23\} layers and then double the layers to \{7, 11, 15, 19, 23, 27\}. This process is same for other LLMs with different layers. The default RoI number $ n $ is set to 4, and the multi-scale HR Resolution $m, \lambda$ is set to 3 and 2/3/4$\times$, respectively.

\begin{figure}[h]
    \centering
    \includegraphics[width=1.0\linewidth]{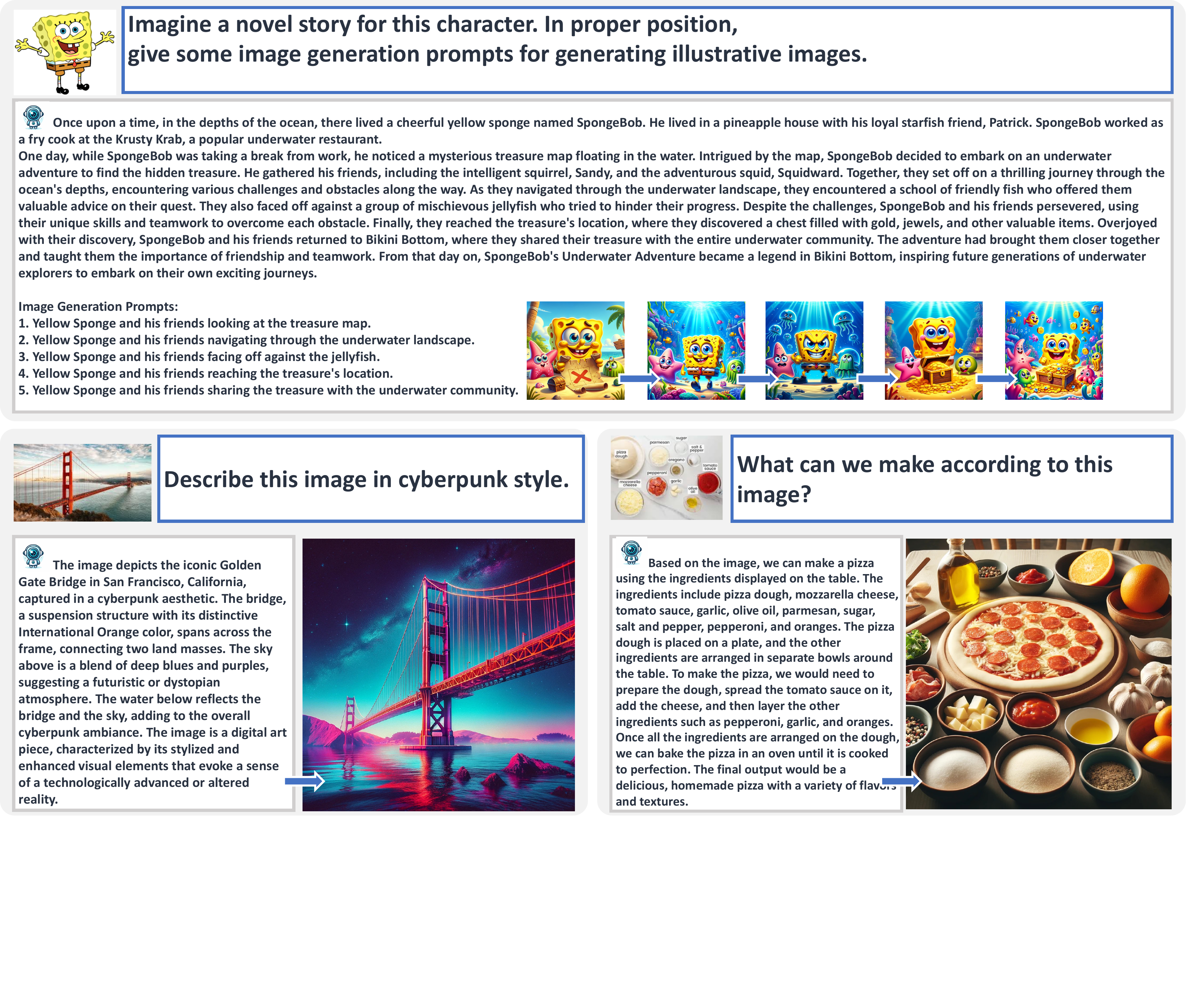}
    \vspace{-25mm}
    \caption{Reasoning Generative Capability. The generated images are from DALLE3 using prompts generated by \modelname.}
    \label{fig:demo_generative}
\end{figure}

\paragraph{Reasoning Generative Ability}
In Figure~\ref{fig:demo_generative}, We also demonstrate \modelname's ability in generative reasoning, which means that when given an image, \modelname can firstly reason from it then produce text with image-generation prompts based on instructions. In Sponge-Bob example, \modelname is capable of generating a story accompanied by multiple illustrative images. Also, in the Golden Gate Bridge scenario, \modelname can edit the source image in various ways. In the pizza example, \modelname can analyze the image and given ingredients to produce an article and illustrative image for making the pizza.

\newpage

\end{document}